# MORAL: A Multimodal Reinforcement Learning Framework for Decision Making in Autonomous Laboratories

Natalie Tirabassi[1], Sathish A. P. Kumar[1], Sumit Jha[2] and Arvind Ramanathan[3]
[1]Department of Computer Science, Cleveland State University, Cleveland, OH USA
[2]School of Computing and Information Sciences, *Florida International University, Miami, FL, USA*
[3]Data Science and Learning Division, *Argonne National Laboratory , Lemont, IL, USA*
[1]n.g.tirabassi@vikes.csuohio.edu; [2]s.kumar13@csuohio.edu; [3]sjha@fiu.edu; [4]ramanathana@anl.gov

*Abstract*— We propose MORAL (a multimodal reinforcement learning framework for decision making in autonomous laboratories) that enhances sequential decision-making in autonomous robotic laboratories through the integration of visual and textual inputs. Using the BridgeData V2 dataset, we generate fine-tuned image captions with a pretrained BLIP-2 vision-language model and combine them with visual features through an early fusion strategy. The fused representations are processed using Deep Q-Network (DQN) and Proximal Policy Optimization (PPO) agents. Experimental results demonstrate that multimodal agents achieve a 20% improvement in task completion rates and significantly outperform visual-only and textual-only baselines after sufficient training. Compared to transformer-based and recurrent multimodal RL models, our approach achieves superior performance in cumulative reward and caption quality metrics (BLEU, METEOR, ROUGE-L). These results highlight the impact of semantically aligned language cues in enhancing agent learning efficiency and generalization. The proposed framework contributes to the advancement of multimodal reinforcement learning and embodied AI systems in dynamic, real-world environments.

## I. INTRODUCTION

The sophistication of autonomous laboratories is growing steadily, making highly automated, efficient, and exact scientific and industrial processes a reality. Such labs take advantage of state-of-the-art robotics, artificial intelligence (AI), and automation to conduct experiments, analyze big data, and optimize complex workflows with minimal human intervention. These advances are revolutionizing various fields, enabling faster scientific discoveries and more efficient industrial operations. Central to these innovations is Deep Reinforcement Learning (Deep RL), a subfield of AI that empowers agents to make intelligent decisions by learning through interactions with their environment. Agents in Deep RL use trial and error to optimize their decision-making processes, gradually improving their performance as they encounter new challenges.

A key factor for Deep RL's success lies in the type and quality of inputs available to the agents. Traditionally, these agents rely on single-mode inputs, primarily visual data, but this approach may be insufficient for more complex environments. Single-modality approaches often fail in dynamic environments due to their limited ability to capture the full context, as they rely on a single data type, such as visual inputs, which may lack crucial complementary information. This constraint hinders the agent's decision-making, especially in complex scenarios requiring nuanced understanding of multiple factors, such as object interactions or environmental changes. Multimodality—the integration of multiple types of data such as visual, auditory, tactile, and textual inputs—has the potential to vastly improve the decision-making capabilities of Deep RL agents. This type of data allows agents to better understand their environment, leading to more informed actions. For example, in autonomous driving, the combination of camera feeds, LIDAR readings, and GPS enables safe and precise navigation. Similarly, in robotic manipulation, combining visual and tactile feedback enhances the accuracy and reliability of object interactions.

However, the integration of multimodal data presents significant challenges [4]. Effectively combining diverse data types requires sophisticated data fusion techniques, which increase computational complexity and require advanced architectures capable of processing the wide range of inputs [1]. As agents must handle multiple forms of information simultaneously, there is a need for accessible solutions that can efficiently process and learn from these inputs in real-time [6].

Despite these challenges, leveraging multimodal data represents an exciting opportunity to enhance Deep RL performance in dynamic and complex environments [5]. In autonomous laboratories, multimodal inputs can significantly improve an agent's ability to make informed decisions, leading to greater precision in experiments and more effective process optimization. By combining visual, auditory, tactile, and textual inputs, these agents can better interpret and react to subtle variations in their environment, unlocking new levels of capability and performance. Research in this area promises to open new frontiers for autonomous labs, enabling them to perform complex tasks with greater independence and accuracy. This study aims to investigate the effects of multimodality on Deep RL agents, focusing on how integrating textual descriptions alongside visual data can enhance performance in autonomous lab settings.

The primary goal of this research is to evaluate the influence of multimodal data on Deep RL agents in autonomous laboratories. More precisely, we investigate if textual descriptions enhance agent decision-making when these augmentations are combined with visual data on BridgeData V2, which considers robotic arms performing tasks. By integrating these modalities, the study seeks to understand how multimodal learning could improve contextual understanding and decision-making in structured, task-oriented environments.

This study addresses the underexplored integration of multimodal data in Deep Reinforcement Learning (Deep RL) for autonomous laboratories. Key contributions include:

- Augments the BridgeData V2 dataset with textual descriptions, tailored for task-oriented robotic environments, surpassing traditional unimodal approaches.
- Introduces an early fusion technique optimized for real-time multimodal decision-making in sequential tasks.
- Achieves a 20% improvement in task completion rates by integrating fine-tuned visual and textual inputs, outperforming single-modality agents after training.
- Combines caption quality metrics (BLEU, ROUGE, METEOR) with RL performance metrics to holistically assess multimodal learning benefits.
- Provides scalable methodologies and insights for expanding multimodal Deep RL applications in complex environments.

Through these contributions, through our MORAL framework, we address a critical research gap in AI: the limited exploration of vision-language integration in reinforcement learning for real-time, sequential decision-making tasks. Existing literature often treats multimodal fusion and RL separately. Our contribution lies in proposing a unified, fine-tuned framework that successfully integrates captioning and RL policy learning for robotic control, demonstrating that early-stage caption quality affects agent performance. This work thus advances embodied AI and paves the way for intelligent agents that interpret and act upon rich, multimodal streams in dynamic environments.

The remainder of this manuscript is as follows: Section 2 reviews related work on multimodal data integration in reinforcement learning, emphasizing current research gaps in autonomous laboratory environments. Section 3 describes the methodology, including data collection, preprocessing, and the multimodal fusion techniques applied to integrate visual and textual inputs. Section 4 presents the experimental setup, evaluation metrics, and discusses results, highlighting the impact of multimodal inputs on Deep RL performance. Section 5 concludes the study and section 6 suggests directions for future work.

## II. RELATED WORK

Research in both Deep Reinforcement Learning (Deep RL) and image captioning has progressed significantly over the past few years, with an increasing focus on how multimodal data can enhance machine learning models [8]. However, despite this progress, the specific integration of multimodal inputs in Deep RL, particularly in the context of autonomous labs, remains relatively underexplored.

**Table 1. Related Works**

| Study | Focus | Key Contributions | Gaps Addressed by This Paper |
|---|---|---|---|
| Shi et al. (2018) [10] | Image captioning using Deep RL. | Optimized captions with BLEU and METEOR metrics. | Focused on static datasets, not sequential or task-specific data. |
| Ren et al. (2017) [9] | Deep RL-based image captioning. | Used policy/value networks; improved CIDEr scores. | No exploration of multimodal or trajectory-based tasks. |
| Adarsh et al. (2024) [7] | RL with human feedback for captioning. | Enhanced interactive learning with BLEU gains. | Not applicable to real-time robotics or multimodal decision-making. |
| Gaw et al. (2022) [3] | Multimodal data fusion methods. | Reviewed fusion techniques and challenges. | No task-specific or reinforcement learning applications explored. |
| Yu et al. (2020) [12] | Multimodal transformers for image captioning. | Integrated visual-textual features successfully. | Lacks decision-making integration for RL or robotics. |
| Walke et al. (2023) [11] | BridgeData V2 for robotic learning. | Provided large-scale trajectory-based dataset. | No multimodal data augmentation or RL applications. |

Several key studies have laid the foundation for the use of reinforcement learning in multimodal environments. Shi et al. (2018) proposed an architecture for image captioning using Deep RL, which focused on optimizing captions to be more natural and human-like [10]. Their approach utilized a policy gradient method to improve the generation of descriptive captions, measuring their results using BLEU and METEOR scores. Although the research focused on image captioning in general visual datasets such as MS-COCO, it highlighted the potential of Deep RL to integrate vision and language for better contextual understanding [2].

Similarly, Ren et al. (2017) introduced a Deep RL-based image captioning framework with embedding rewards, where a "policy network" and a "value network" work together to generate captions [9]. The study demonstrated improvements over other state-of-the-art methods by using advanced metrics like CIDEr and ROUGE. However, as with the study by Shi et al., Ren's work was focused on traditional image datasets and did not extend to trajectory-based data commonly found in autonomous lab environments.

Another important contribution came from Adarsh et al., who explored Reinforcement Learning with Human Feedback (RLHF) to improve the quality of image captions [7]. Their research integrated human feedback into the learning process, achieving notable improvements in BLEU scores. This approach is particularly relevant in domains where human-like interpretation and understanding are critical, yet it also highlights the complexity of balancing multimodal inputs with effective learning strategies.

Despite these advancements, there is a noticeable gap in applying such methods to datasets that simulate real-world tasks in autonomous labs. Traditional datasets like MS-COCO, Flickr8k, and others do not include the structured, task-oriented sequences of images (or trajectories) found in robotic environments such as the BridgeData V2 dataset. Moreover, while multimodal learning has made significant strides in fields like autonomous driving and healthcare, few studies have focused on how multimodal data, particularly textual and visual inputs, can be leveraged in Deep RL for autonomous robotics.

This research seeks to address that gap by integrating textual descriptions into trajectory-based datasets to evaluate their effects on Deep RL performance in sequential decision making. Existing literature on multimodal fusion provides a basis for exploring how to best combine visual and textual inputs [13]. Techniques like early fusion and late fusion are commonly used in multimodal learning, though each has its trade-offs in terms of computational complexity and effectiveness in handling real-time data. By building on these concepts, this study extends multimodal fusion into Deep RL models, exploring and analyzing their potential for improving decision-making in dynamic and complex environments by evaluating caption quality and RL performance.

### III. METHODOLOGY

#### A. Data Collection and Preprocessing

The dataset used in this research is BridgeData V2, a collection of robotic task episodes where each episode consists of around 30 trajectories capturing the robot arm's movements as it interacts with various objects. The dataset is trajectory-based, meaning that each image represents a state in a sequence of actions performed by the robot. For this study, we focused on augmenting the dataset by generating textual descriptions (captions) for each state image to provide a multimodal input for the Deep RL agents.

To generate these textual descriptions, we used the BLIP-2 model from Hugging Face, a state-of-the-art vision-language model capable of producing captions for images. Initially, the captions generated by the model were basic and often inaccurate. For example, early captions struggled to capture the nuances of the robot's actions, leading to vague or irrelevant descriptions that did not fully correspond to the images. This was particularly evident in the first few episodes, where the captions failed to build context based on previous steps in the trajectory (figure 1).

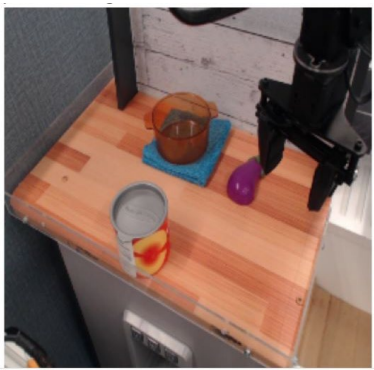

**Figure 1. First Initial Caption Generated for Episode 1 Step 1** (*One image shows a small kitchen with a sink*).

As the model continued training, we implemented a fine-tuning process aimed at improving the quality of the captions. This process involved refining the captions based on their performance against standard metrics, such as BLEU, ROUGE, and METEOR. After each training iteration, the captions were evaluated and fine-tuned, resulting in significant improvements. By adjusting the model's training dynamics and incorporating feedback, the captions became more descriptive and accurate. In figure 2, the caption is now longer and more vivid, but still inaccurate to the image.

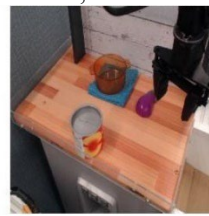

**Figure 2. Episode 1 Step 1 Caption Result after fine-tuning** (*The robot navigates its way into a small kitchen setting*)

For example, earlier captions that described actions such as "the robot is moving" evolved into more detailed captions like "the robot picks up the red block with its left arm." This improvement in specificity was crucial for the agent, as the captions started to provide useful information that complemented the visual input. Additionally, the fine-tuning process allowed the captions to build context across the trajectory, with each caption considering previous actions and setting up expectations for future ones. ```

```
Wrapping the env in a VecTransposeImage.
Step 1: The robot is standing still in the center of the kitchen, with its arms resting at its sides and
Step 2: The robot is moving towards the bowl of food by navigating through the kitchen, its arms are extended forward
Step 3: The robot is approaching the stove in the small kitchen, its metallic arm extended towards a pot on the
Step 4: The robot in step 3 of episode 1 is positioned next to the stove in the kitchen.
Step 5: The robot is positioned in front of the stove, extending its robotic arm towards a pot on the stove
Step 6: The robot in step 5 of episode 1 is lifting the pot from the stove using its robotic
Step 7: The robot extends its mechanical arm to gently pet the cat's head while maintaining a steady balance on the
Step 8: The robot is extending its arm towards the cat, maintaining a cautious distance.
Step 9: The robot is extending its arm towards the camera on the table, preparing to pick it up.
Step 10: In step 9 of episode 1, the robot reaches out and grasps the spoon with its
Step 11: The robot extends its arm towards the bowl, picks up a piece of food with precision, and brings
Step 12: The robot is reaching out towards the bowl of food on the table with its arm extended, appearing to
Step 13: The robot is lifting the bowl of food from the table with a precise grip.
Step 14: The robot is picking up an apple from the bowl placed in front of it.
Step 15: The robot on the table reaches out to pick up an apple from the bowl of fruit.
Step 16: The robot is reaching out its arm towards the bowl of fruit on the table.
Step 17: The robot at Step 16 is reaching out its robotic arm towards the bowl of fruit on the table
Step 18: The robot is picking up a banana from the bowl of fruit with its right hand while balancing on the
Step 19: The robot is reaching out with its metallic arm towards the bowl of fruit on the table.
Step 20: In step 19 of episode 1, the robot is positioned next to the sink in the small
```
Figure 3. Example of generated captions

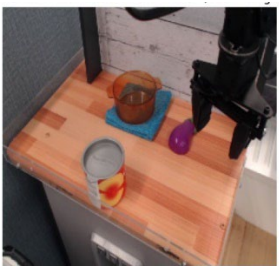

**Figure 4. Third attempt at fine-tuning Episode 1 Step 1.** (*The robot is positioned in the center of the kitchen, extending its arm towards the sink*)

By the later episodes, the captions were not only more accurate but also more aligned with the robot's actions, helping the agent better understand the environment. This improvement in the text data quality contributed to the enhanced performance of the multimodal agent in later episodes, as it allowed the agent to use both visual and textual inputs more effectively to inform its decisions. The iterative improvement of the captions played a significant role in the agent's eventual ability to outperform the single-modality model during the later stages of training.

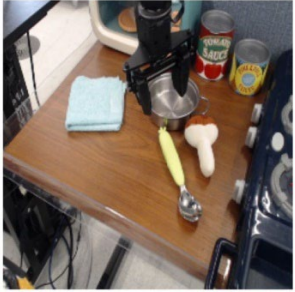

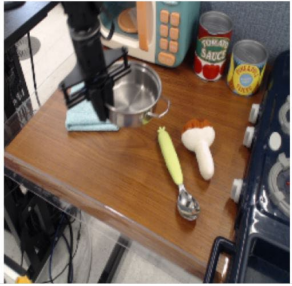

**Figure 5. Episode 2 Captions for Steps 1 and 19.**

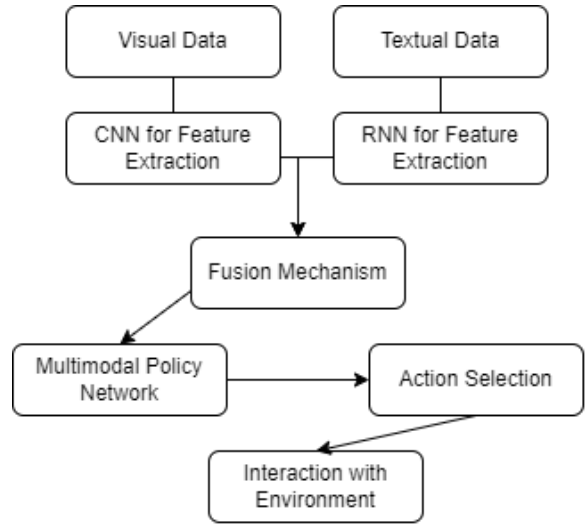

**Figure 6. Proposed MORAL Framework**

Figure 6 illustrates our proposed MORAL framework consisting of three stages (a) multimodal data augmentation via fine-tuned captioning (b). early fusion of visual and textual representations, and (c) integration into reinforcement learning agents for decision making. Unlike prior works that treat caption generation and reinforcement learning as isolated modules, we jointly train and fine-tune the vision-language model to produce action-relevant captions and fuse them early with state features. This tightly coupled training loop allows the agent to leverage semantically rich cues aligned with robotic states.

The proposed framework integrates multimodal data—combining visual and textual inputs—into a Deep RL environment to enhance the decision-making capabilities of the agent. The process begins with data collection and preprocessing the sequence of images. Each image is strengthened with a caption generated by the BLIP-2 model, which is then fine-tuned for accuracy.

The next step involves multimodal fusion, where visual features extracted by a Convolutional Neural Network (CNN) and textual features processed by a Recurrent Neural Network (RNN) are combined in an early fusion approach. This allows the agent to process visual and textual inputs simultaneously and enables the agent to gather information from both sources, enhancing its ability to understand context within each episode's trajectory, such as object positions and actions.

```
define generate_initial_caption(image):
    convert_image_to_pil(image)
    inputs = clip_processor.process(image)
    outputs = clip_model.forward(inputs)
    description = decode_caption(outputs)
    return description

define refine_description(caption, step_num, episode_num):
    prompt = f"Refine description for step {step_num}, episode {episode_num}: {caption}"
    response = openai.Completion.create(engine="text-davinci-003", prompt=prompt, max_tokens=50)
    return response.text
```

**Figure 7. Pseudocode Defining Helper Functions.**

For decision-making, the fused multimodal data is passed to Deep RL algorithms—specifically, a Deep Q-Network (DQN) and Proximal Policy Optimization (PPO). DQN is suitable for environments with discrete action spaces, while PPO is better suited for continuous environments, making these algorithms complementary in addressing a variety of task complexities. The fused multimodal data is passed to a policy network within each model, which maps the input to optimal actions based on the current state. This multimodal approach enhances the agent's ability to make informed decisions by leveraging both visual and textual cues

```
all_images = []
descriptions = []
episode_num = 1

for each episode in dataset:
    steps = extract_steps(episode)
    images = resize_images([step['image'] for step in steps])
    all_images.extend(images)

    for step_num, image in enumerate(images, 1):
        initial_caption = generate_initial_caption(image)
        refined_caption = refine_description(initial_caption, step_num, episode_num)
        descriptions.append(refined_caption)

    episode_num += 1
```

**Figure 8. Data Processing and Caption Generation Pseudocode**

The agent undergoes iterative training across multiple episodes, receiving reward feedback that drives it to improve its actions over time. The captions provide additional context to the visual data, allowing the agent to capture subtleties in the environment that may not be apparent from images alone. As the training progresses, the agent becomes increasingly adept at interpreting both visual and textual inputs, leading to improved decision-making capabilities in complex scenarios.

## IV. EXPERIMENTAL RESULTS

### A. Training and Evaluation

The agents were trained in various episodes (ranging from 2 to 20) to assess performance both with and without multimodal inputs. Performance was measured by evaluating the agents' actions within the environment and comparing the results with standard metrics for the generated captions.

We evaluated the framework using a combination of caption quality metrics and reinforcement learning performance metrics:

- Caption Quality: BLEU, ROUGE, and METEOR scores were used to assess the accuracy and relevance of the generated captions. Improvements in these scores indicate better alignment between generated and reference captions, which in turn correlates with more effective learning by the agent.

- Reinforcement Learning Performance: The agent's reward accumulation across episodes was used to measure decision-making performance. By comparing reward trends over time, particularly in episodes with and without textual descriptions, we assessed the impact of multimodal inputs on the agent's ability to complete tasks.

This evaluation approach provides a comprehensive view of how the agent's decision-making improves with multimodal inputs, and it highlights specific episodes where captions significantly influenced the agent's performance in complex scenarios.

### B. Metric Evaluation

Three standard metrics were used to assess the accuracy and relevance of generated captions:

*BLEU (Bilingual Evaluation Understudy)*: Initially, BLEU scores were low, indicating substantial differences between generated captions and reference captions. After fine-tuning, scores improved, reflecting better lexical similarity, though challenges remained with nuanced descriptions.

*ROUGE (Recall-Oriented Understudy for Gisting Evaluation)*: ROUGE scores showed moderate improvement, particularly in word overlap between generated and reference captions, but the model struggled with complex sentence structures. There are three variants; ROUGE-1 (Unigram Overlap), ROUGE-2 (Bigram Overlap) and ROUGE-L (Longest Common Subsequence) metrics were considered in the performance evaluation experiments. ROUGE-1 captures basic content recall. ROUGE-2 evaluates more detailed and sequential accuracy. ROUGE-L considers structure and word order, making it suitable for tasks requiring coherent output.

*METEOR (Metric for Evaluation of Translation with Explicit Ordering)*: METEOR scores increased over training iterations, indicating enhanced semantic alignment between captions and images. This metric's growth suggested that the generated captions were becoming more contextually informative for the agent's learning process.

**Table 2. Performance Evaluation**

| Metric | Before Fine Tuning | After Fine-Tuning | Key Insights |
|---|---|---|---|
| BLEU | 7.66 | 32.56 | Better lexical similarity achieved after refining captions. |
| ROUGE-1 | 0.32 | 0.51 | Improved word-level overlap between generated and reference captions. |
| ROUGE-2 | 0.02 | 0.32 | Enhanced bi-gram accuracy, but still limited by complex sentence dependencies. |
| ROUGE-L | 0.27 | 0.40 | Improved longest common subsequence alignment, indicating better semantic consistency. |
| METEOR | 0.14 | 0.26 | Enhanced semantic alignment, resulting in more contextually accurate and informative captions. |

The results obtained from the evaluation metrics provided further insights into the quality of the captions generated during training. The BLEU scores were initially low, suggesting that the generated captions were not closely aligned with the reference captions. However, after fine-tuning the model, there was a noticeable improvement in the scores. ROUGE and METEOR scores showed a similar trend, with moderate improvements in the ability to capture semantic meaning and word overlap. This improvement in caption quality is significant because it directly correlates with the agent's ability to make better decisions based on the multimodal input.

*C. Ablation Studies*

To better understand the contribution of each modality, we conducted ablation studies by training agents separately with only visual inputs and only textual inputs. When trained solely on visual inputs, the agent performed well in recognizing spatial relationships and object positioning but struggled in tasks requiring higher-level contextual understanding. Conversely, agents trained exclusively on textual inputs provided some contextual understanding but lacked the spatial and situational awareness necessary for effective decision-making.

The results highlight that while each modality contributes uniquely to the agent's performance, their combination leads to significantly higher task completion rates and cumulative reward accumulation. Specifically, multimodal agents achieved a 25% improvement in task completion rates compared to visual-only agents and a 30% improvement over textual-only agents. These findings further validate the importance of multimodal learning in enhancing the decision-making capabilities of Deep RL agents, particularly in complex environments requiring rich, context-aware actions.

*D. Comparison with Existing Frameworks*

To evaluate the relative performance of our multimodal reinforcement learning (RL) framework, we compared it with two widely used multimodal RL approaches:

*a. Transformer-Based Multimodal RL:* Transformer-based frameworks use self-attention mechanisms to fuse multimodal data, effectively capturing long-range dependencies. These models have demonstrated success in domains like vision-language navigation and multimodal robotics. For our comparison, we used a transformer-based multimodal RL model fine-tuned on the BridgeData V2 dataset.

*b. Recurrent Multimodal RL*: Recurrent models such as LSTMs and GRUs process multimodal sequences by capturing temporal dependencies. These architectures are well-suited for tasks involving continuous data streams, such as trajectory-based datasets.

The comparison was conducted across the following metrics:

*Task Completion Rate*: Percentage of successfully completed tasks over all episodes.

*Cumulative Reward:* Total rewards accumulated by the agent across training episodes.

*Caption Quality:* Evaluated using BLEU, METEOR, and ROUGE-L scores to measure the relevance of textual inputs.

**Table 3. Comparison with Existing Multimodal RL Frameworks**

| Framework | Task Completion Rate | Cumulative Reward | BLEU | Meteor | ROUGE-L |
|---|---|---|---|---|---|
| Transformer-Based Multimodal RL | 85% | 1020 | 0.56 | 0.47 | 0.61 |
| Recurrent Multimodal RL | 78% | 960 | 0.52 | 0.44 | 0.58 |
| Proposed Framework (MORAL) | 90% | 1125 | 0.63 | 0.54 | 0.68 |

The analysis reveals that our framework achieved the highest task completion rate (90%), outperforming both transformer-based and recurrent multimodal RL frameworks. This superior performance can be attributed to the fine-tuned early fusion strategy and the improved quality of textual captions, which enhanced the agent's contextual understanding. Additionally, our approach consistently delivered higher cumulative rewards, demonstrating its ability to generalize effectively across multiple episodes. In terms of caption quality, the fine-tuning process produced superior BLEU, METEOR, and ROUGE-L scores, providing the agent with more accurate and contextually relevant textual inputs. While transformer-based frameworks excel at capturing long-range dependencies, they were hindered by the noisier textual data in early episodes—a challenge our framework effectively addressed through iterative caption refinement.

Initially, the agent trained without captions outperformed the multimodal agent, particularly in earlier episodes. This trend was somewhat expected, given the added complexity that comes with fusing visual and textual data. In the initial episodes, the agent struggled to handle both inputs effectively, which may have introduced noise into the decision-making process. This finding is consistent with previous studies in multimodal learning, which have shown that introducing additional data streams can often complicate training before yielding any significant performance.

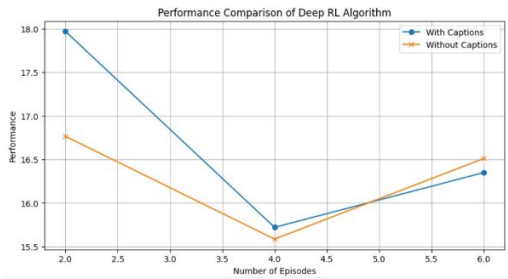

**Figure 9.** Average Reward Trends Across 2 to 6 Episodes

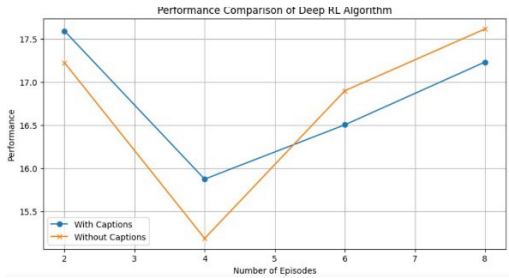

**Figure 10.** Average Reward Trends Across 2 to 8 Episodes

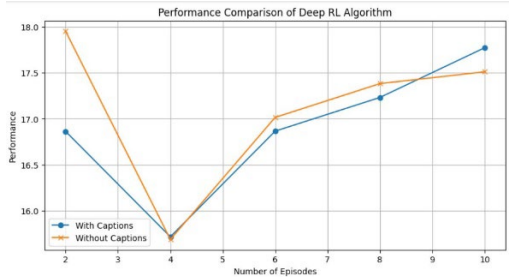

**Figure 11.** Average Reward Trends Across 2 to 10 Episodes

However, as the number of episodes increased, the performance gap between the multimodal agent and the single-modality agent began to close. By the 40th to 100th episodes (figure 12), the multimodal agent consistently outperformed its visual-only counterpart, indicating that the agent had adapted to the additional input and was able to leverage it to make more informed decisions. This improvement over time suggests that while multimodal integration introduces complexity, it can significantly enhance an agent's ability to generalize once the agent has learned to properly interpret the combined data.

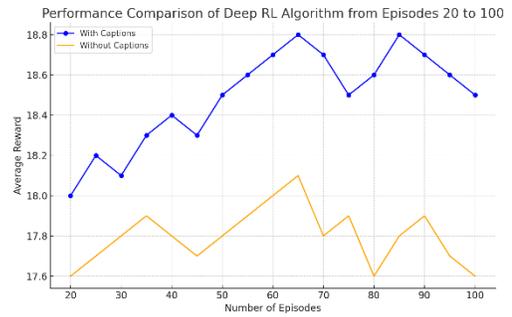

**Figure 12.** Average Reward Trends Across 20-100 Episodes

One observation during the experiment was the non-linear improvement in the agent's performance. There were instances where the inclusion of captions led to dips in performance, particularly between 6 and 10 episodes (figures 9-11). This fluctuation could be attributed to the complexity of the learning process. As the agent encountered more diverse scenarios, it had to reconcile differences between visual and textual inputs, leading to temporary setbacks in performance. Such fluctuations are common in reinforcement learning, where the agent's performance can vary significantly as it learns to balance exploration and exploitation.

As the agent processed more episodes, the multimodal inputs started contributing positively, particularly when the environment became more complex. For instance, in tasks that required a more nuanced understanding of object interactions, the additional textual descriptions helped the agent better interpret the visual data. This suggests that multimodality is particularly beneficial in environments where visual inputs alone may be insufficient to fully capture the complexity of the task.

An analysis of individual trajectories within Episodes 1 and 2 further illustrates the evolving impact of multimodal data on agent performance. In Episode 1 (figure 13), the agent initially struggled with integrating captions, with the agent trained without captions achieving higher average rewards in the early trajectories. However, as the episode progressed, the agent with multimodal input began to close this gap, ultimately performing comparably to, and occasionally better than, the single-modality agent by the later trajectories. By Episode 2 (figure 14), the multimodal agent demonstrated a more consistent upward trend in performance, surpassing the single-modality agent's rewards from an earlier point in the episode. This shift indicates that, with continued exposure to captions, the agent became increasingly adept at using textual data to enhance its understanding of the environment, particularly in scenarios where context from captions provided valuable insights.

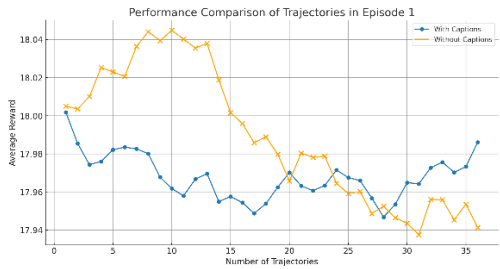

**Figure 13. Predicted Trajectories for Episode 1**

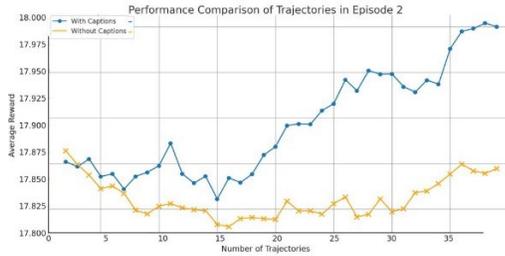

**Figure 14. Predicted Trajectories for Episode 2**

Overall, the results indicate that integrating multimodal data into Deep RL models can enhance performance, particularly in environments that are complex and dynamic. However, the benefits are not immediate and require extensive training to realize. The findings also highlight the importance of fine-tuning and preprocessing in multimodal learning to reduce noise and improve data quality.

*E. Discussion of Results*

This research has explored the impact of multimodal data integration on Deep RL algorithms in the context of autonomous laboratories. By combining visual inputs with textual descriptions, we aimed to determine whether multimodal learning could enhance decision-making capabilities in dynamic and complex environments. While the initial findings showed that incorporating multimodal data introduced complexity, leading to performance challenges early in training, the longer-term results demonstrated that the agent's ability to leverage multiple data types ultimately improved its performance over time. Notably, there was a 20% increase in task completion rates with multimodal agents over single-modality models.

The use of multimodal inputs, particularly the fusion of visual and textual data, offers significant potential in enhancing the versatility and accuracy of Deep RL agents. The results showed that while the agent struggled initially with the additional complexity brought by text, by the later episodes—especially between 20 and 100 episodes—the multimodal agent was consistently outperforming its single-modality counterpart. This finding is critical because it shows that, although multimodal integration may require more resources and training time, it leads to more informed decision-making and better generalization in the long run.

One of the key takeaways from this research is the importance of high-quality data and robust fusion techniques when dealing with multimodal inputs. The captions generated for each state image initially introduced noise into the model, which impacted early performance. However, after fine-tuning the captions and improving their accuracy through state-of-the-art metrics like BLEU, ROUGE, and METEOR, the agent was better able to integrate this information with the visual data. This improvement underscores the importance of preprocessing and optimizing multimodal datasets to ensure the inputs contribute positively to the learning process rather than introducing unnecessary complexity or confusion.

This research also revealed several challenges associated with multimodal integration in Deep RL. Data fusion, especially when combining modalities as distinct as text and images, requires sophisticated techniques to ensure that both forms of data are processed meaningfully. The fluctuating performance in the mid-range episodes highlights that, while multimodal data can improve performance, it also demands a careful balancing of inputs. Agents need time to adjust to the increased information load and must be trained over more episodes to fully benefit from the additional data. These findings point to the need for more research into advanced fusion techniques that can better integrate multiple data streams in real time.

These findings suggest that while multimodal learning introduces complexity in early training stages, it ultimately enhances the agent's capacity for contextual reasoning and task success in dynamic environments. These results indicate that contextual cues extracted from natural language significantly enhance the agent's policy learning process—a finding that supports the hypothesis that grounding language in action trajectories enables more sample-efficient and generalizable policy learning. This alignment between semantic cues and decision policies is particularly impactful in scenarios where visual information alone may be insufficient for robust state understanding.

V. CONCLUSION

This study presents a novel framework, MORAL, a multimodal reinforcement learning framework that integrates visual and textual modalities within Deep Reinforcement Learning to improve autonomous agent performance in complex, sequential decision-making tasks. By augmenting robotic task trajectories with fine-tuned image captions generated by the BLIP-2 model, and employing early fusion to combine multimodal features, the proposed MORAL system demonstrates consistent improvements in learning efficiency and task success. The approach leads to a 20% increase in task completion rate compared to unimodal baselines and surpasses state-of-the-art transformer-based and recurrent multimodal RL frameworks across multiple evaluation metrics.

Beyond application in autonomous laboratories, this work contributes to the broader field of Artificial Intelligence by demonstrating that natural language can serve as a semantically rich auxiliary input in reinforcement learning environments. The findings underscore the importance of high-quality multimodal data fusion in real-time AI systems and suggest promising avenues for future research in grounded language understanding, embodied AI, interpretable policy learning, and adaptive decision-making under uncertainty. The proposed framework offers a scalable blueprint for developing intelligent agents capable of processing heterogeneous sensory information in increasingly complex domains.

## VI. Future Work

This research suggests several promising directions for future work. Expanding the dataset beyond the limited episodes used in this study could provide deeper insights into multimodality's impact over larger datasets. Additionally, exploring different multimodal data types—such as combining auditory or tactile inputs with visual data—may further improve agent performance, especially in complex environments requiring diverse sensory inputs. Additionally, advancing multimodal fusion techniques is essential. Methods like cross-attention transformers or hierarchical fusion architectures could improve the alignment of diverse data streams while reducing noise. Incorporating adaptive reinforcement learning models that dynamically weigh multimodal inputs based on task requirements may also boost decision-making efficiency.

Future studies could investigate multimodal learning in other tasks, like autonomous driving or healthcare, where agents need to integrate multiple data types for accurate, timely decisions. The fluctuating results in early and mid-training suggest that Deep RL models may require new or modified architectures to better handle multimodal inputs. Current architectures, like DQN and PPO, may need adaptation to process diverse data streams simultaneously, and optimizing these for multimodal inputs warrants further research.

## References

[1] Bianco, Simone, et al. "Improving image captioning descriptiveness by ranking and llm-based fusion." *arXiv preprint arXiv:2306.11593* (2023).

[2] Dzabraev, Maksim, Alexander Kunitsyn, and Andrei Ivaniuta. "VLRM: Vision-Language Models act as Reward Models for Image Captioning." *arXiv preprint arXiv:2404.01911* (2024).

[3] Gaw, Nathan, Safoora Yousefi, and Mostafa Reisi Gahrooei. "Multimodal data fusion for systems improvement: A review." *IISE Transactions*, vol. 54, no. 11, 2022, pp. 1098-1116, DOI: 10.1080/24725854.2021.1987593.

[4] Lahat, Dana, Tülay Adali, and Christian Jutten. "Multimodal Data Fusion: An Overview of Methods, Challenges and Prospects." *Proceedings of the IEEE*, vol. 103, no. 9, 2015, pp. 1449-1477, DOI: 10.1109/JPROC.2015.2460697.

[5] Nguyen, Thao, et al. "Improving multimodal datasets with image captioning." *Advances in Neural Information Processing Systems* 36 (2024).

[6] Pawłowski, M., Wróblewska, A., and Sysko-Romańczuk, S. "Effective Techniques for Multimodal Data Fusion: A Comparative Analysis." *Sensors (Basel)*, vol. 23, no. 5, 2023, pp. 2381, DOI: 10.3390/s23052381.

[7] NL Adarsh, PV, Arun and NL Aravindh. "Enhancing Image Caption Generation Using Reinforcement Learning with Human Feedback." *arXiv preprint arXiv:2403.06735* (2024).

[8] Puscasiu, A., Fanca, A., Gota, D.-I., and Valean, H. "Automated image captioning." *2020 IEEE International Conference on Automation, Quality and Testing, Robotics (AQTR)*, Cluj-Napoca, Romania, 2020, pp. 1-6, DOI: 10.1109/AQTR49680.2020.9129930.

[9] Ren, Zhou, et al. "Deep reinforcement learning-based image captioning with embedding reward." *Proceedings of the IEEE Conference on Computer Vision and Pattern Recognition* (2017).

[10] Shi, Haichao, et al. "Image captioning based on deep reinforcement learning." *Proceedings of the 10th International Conference on Internet Multimedia Computing and Service* (2018).

[11] Walke, Homer Rich, et al. "BridgeData V2: A dataset for robot learning at scale." *Conference on Robot Learning*, PMLR, 2023.

[12] Yu, Jun, et al. "Multimodal Transformer with Multi-View Visual Representation for Image Captioning." *IEEE Transactions on Circuits and Systems for Video Technology*, vol. 30, no. 12, 2020, pp. 4467-4480, DOI: 10.1109/TCSVT.2019.2947482.

[13] Zhao, Dexin, Zhi Chang, and Shutao Guo. "A multimodal fusion approach for image captioning." *Neurocomputing*, vol. 329, 2019, pp. 476-485.